\documentclass{article}
\usepackage{timer_m1_arxiv,times}
\usepackage[T1]{fontenc}
\usepackage{amsmath,amssymb}
\usepackage{graphicx,booktabs,multirow,makecell,array}
\usepackage{xcolor}
\newcommand{\secondres}[1]{{\underline{\textcolor{blue}{#1}}}}
\usepackage{placeins,wrapfig}
\usepackage{needspace}
\usepackage{hyperref}
\hypersetup{hidelinks,breaklinks=true}
\usepackage{xurl}
\title{Timer-M1: A Multivariate Time Series \\ Foundation Model via Learning Primitives}
\author{Haoran Zhang\textsuperscript{*,1}, Haixuan Liu\textsuperscript{*,1}, Xingjian Su\textsuperscript{*,1}, Yong Liu\textsuperscript{*,2},\\
\textbf{Zhi Chen\textsuperscript{1}, Yuxuan Wang\textsuperscript{1}, Jianmin Wang\textsuperscript{1}, Mingsheng Long\textsuperscript{$\boxtimes$,1}}\\
\normalfont\textsuperscript{1}School of Software, BNRist, Tsinghua University\\
\normalfont\textsuperscript{2}ByteDance, Beijing, China\\
\normalfont\texttt{\{zhang-hr24,liuhaixu21,suxj22\}@mails.tsinghua.edu.cn}\\
\normalfont\textsuperscript{$\boxtimes$}\texttt{mingsheng@tsinghua.edu.cn}}

\date{}
\hypersetup{
  pdftitle={Timer-M1: A Multivariate Time Series Foundation Model via Learning Primitives},
  pdfauthor={Haoran Zhang, Haixuan Liu, Xingjian Su, Yong Liu, Zhi Chen, Yuxuan Wang, Jianmin Wang, Mingsheng Long},
  pdfsubject={Multivariate time series foundation models},
  pdfkeywords={Time Series Forecasting, Foundation Model, Learning Primitives, Zero-Shot Forecasting}
}

\begin{document}
\setlength{\parskip}{4pt}
\maketitle
\begingroup
\renewcommand{\thefootnote}{}
\raggedright
\footnotetext{\textsuperscript{*}Equal Contribution}
\endgroup

\begin{abstract}
We introduce Timer-M1, a pretrained multivariate time series foundation model that learns with primitives for zero-shot forecasting. Across domains, time series share elementary temporal and relational patterns, termed \emph{primitives}, yet differ in how these primitives manifest and evolve across different contexts. Despite progress in zero-shot and task-general forecasting, existing foundation models may still struggle to generalize to complex real-world scenarios. To this end, we develop a primitive-based data synthesis and pretraining pipeline. The synthesis pipeline generates series with temporal primitives shared across domains and then assembles real and generated series into multivariate samples using relational primitives. Afterwards, samples are organized into \emph{episodes} by assigning distinct channel roles as target variates, past-only covariates, and known-future covariates, ensuring that the model is optimized on predictable variates using available exogenous information. Technically, Timer-M1 further adapts gated two-dimensional Transformer blocks that dynamically allocate cross-variate attention across layers. Across three large-scale forecasting benchmarks, Timer-M1 ranks first on both FEV and TIME and second on GIFT-Eval among most recent time series foundation models. These results support effective primitive-based pretraining as a route to robust general forecasting technique across domains and task settings.
\end{abstract}

\section{Introduction}
\label{sec:introduction}

Time series forecasting spans domains and task configurations in real-world applications. Time series foundation models have advanced a shift from task-specific training toward cross-domain pretraining and zero-shot forecasting \citep{das2024decoder,ansari2024chronos,woo2024unified,khwaja2026toto2,liutimer}, demonstrating competitive accuracy against statistical and supervised baselines on multi-domain and multi-task benchmarks \citep{aksu2024gifteval, shchur2025fev, qiao2026time}. Meanwhile, multivariate foundation models \citep{ansari2025chronos, khwaja2026toto2, liu2026falconx} have empowered native capability to jointly predict multiple targets with historical or known-future covariates under observation settings \citep{shchur2025fev, cohen2026time}.

Time series foundation models are targeted to generalize on cross-domain data and deal with heterogeneous input context. 
Previous foundation models have continuously improved toward this goal by task-agnostic training on real-world and synthetic datasets \citep{das2023decoder, dooley2023forecastpfn}. However, real-world time series record complex and noisy dynamics but offer limited control over which patterns are observed; synthetic datasets provide such control but may omit the irregularities of real systems. This complementarity motivates a data construction process that combines observed dynamics with controllable relations, rather than relying on either source alone.

Recurring temporal and relational patterns provide a common basis for learning across diverse domains and forecasting tasks. Classical time series analysis separates trend, seasonal, and remainder components to understand and forecast a series \citep{hyndman2018forecasting}. We refer to such elementary patterns shared across series and domains as \emph{primitives}: temporal primitives describe trends, periodicity, stochastic dependence, and regime changes, while relational primitives describe cross-variate correlation, delayed responses, cointegration, and event effects. These primitives characterize individual regularities rather than specific forecasting scenarios. Just as learned concepts and motifs can be generalized in new combinations in language and vision \citep{zhao2024skillcomposition,liu2022composable}, recurring primitives motivate learning transferable forecasting capabilities that remain useful when their surrounding signals and available observations change across contexts.

\vspace{10pt}
\begin{figure}[t]
\centering
\includegraphics[width=\linewidth]{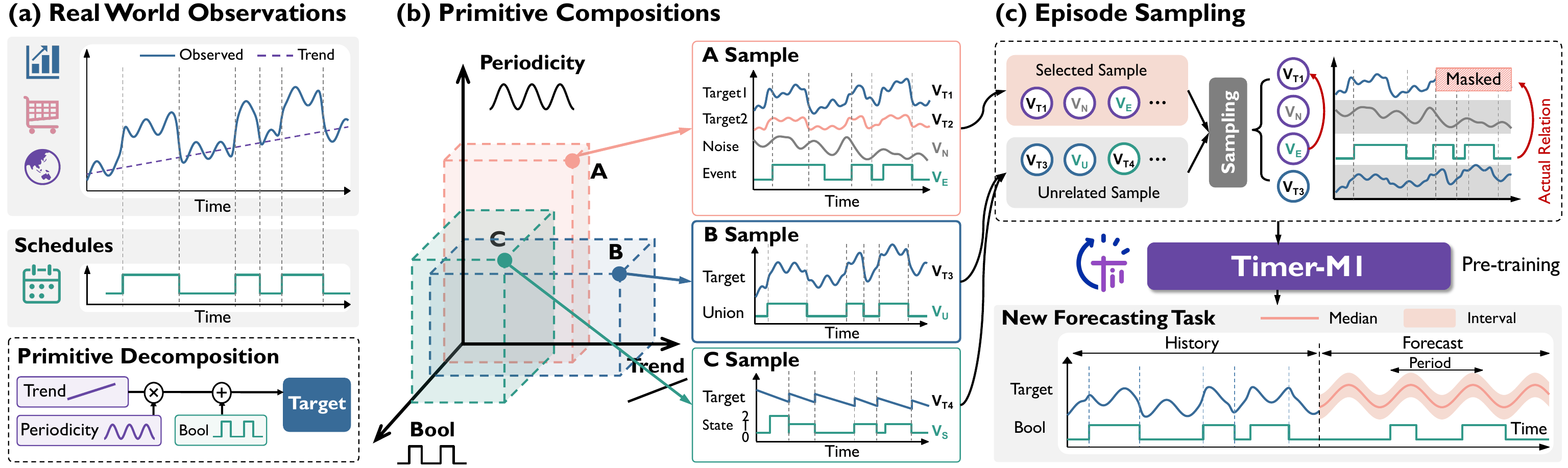}
\caption{Learning with primitives. Temporal and relational patterns observed in real series are composed into diverse samples; episodes vary what is observed and predicted while introducing noise and interference to improve robustness, linking to generalization to new forecasting scenarios.
}
\label{fig:intro}
\end{figure}

Learning with primitives offers a systematic route for training generalizable time series foundation models. As shown in Figure~\ref{fig:intro}, we develop a compact set of operators to describe temporal patterns and cross-variate relations, and to combine synthetic and real-world time series.
In addition to constructing a bottom-up data pipeline, we extend the pretraining process of foundation models into a forecasting task pipeline. We build forecasting episodes that complement metadata of real-world tasks by specifying which variates are targets and which observations are available, coupling relational structure with the information needed for forecasting. Repeated exposure under different roles and masks encourages general forecasting capabilities rather than reliance on a task configuration.

Technically, we build Timer-M1 with native zero-shot and multivariate capabilities via a primitive-based data pipeline and episode-based pretraining. Synthesis operators provide temporal and relational primitives, compose real and generated series into multivariate samples, and apply data augmentations. Episodes assign target, past-only, and known-future roles, via task-dependent supervision masks to guarantee predictability. In terms of the architecture, we adapt gated two-dimensional Transformer blocks to learning from episodes along temporal and variate dimensions, using adaptive gates learned to allocate cross-variate attention across layers. Timer-M1 ranks first on FEV and TIME and second on GIFT-Eval among the state-of-the-art foundation models; controlled primitive analyses further examine the forecasting behaviors underlying these results. Our contributions are:

\begin{itemize}
\item We develop a unified data synthesis pipeline that generates elementary temporal and relational primitives to enrich pretraining data and improve cross-domain generalization.
\item We present Timer-M1, a time series foundation model pretrained on carefully constructed forecasting episodes that unify univariate, covariate-aware, and multivariate forecasting.
\item Timer-M1 achieves strong zero-shot results across tasks, ranking first on FEV and TIME and second on GIFT-Eval, highlighting its potential for broader real-world applications.
\end{itemize}

\section{Related Work}
\label{sec:related}

\subsection{Time Series Pretraining Data}

Pretraining data shape the temporal patterns and cross-variate relations a model encounters. ForecastPFN \citep{dooley2023forecastpfn} learns from synthetic tasks, Chronos \citep{ansari2024chronos} combines real series with Gaussian-process compositions, and Chronos-2 \citep{ansari2025chronos} constructs multivariate dependencies synthetically. Real-data curation remains complementary: TimeBench \citep{liu2025sundial} supplies heterogeneous dynamics, while Timer-S1 \citep{liu2026timers1} combines corpus scaling with transformations to reduce predictive bias. Controlled studies further highlight the importance of data composition and training protocol \citep{wen2026generic}, motivating coordinated data construction and task design for generalizable forecasting.

Learning from generated tasks offers a complementary approach to pretraining data construction. In the tabular domain, TabPFN \citep{hollmann2023tabpfn,hollmann2025tabpfn} uses structural-causal-model priors, TabICL \citep{qu2025tabicl} scales such pretraining to larger contexts, and TabFM \citep{kong2026tabfm} combines hybrid attention with causal-model generation. TabDPT \citep{ma2025tabdpt} instead exploits signals in real tables that synthetic priors may miss. These approaches motivate combining controlled generation with empirical diversity, which we pursue by applying shared relation operators to both real and generated series, composing primitives while retaining observed dynamics.

\subsection{Time Series Foundation Models}

Time series foundation models explore different pretraining objectives and representations for forecasting across domains. Chronos \citep{ansari2024chronos} models discretized values, whereas Sundial \citep{liu2025sundial} develops continuous probabilistic generation. Autoregressive approaches learn transferable temporal patterns through next-token prediction \citep{liutimer,liu2026timers1}, while TiRex \citep{auer2025tirex} adopts contiguous patch masking. Beyond individual series, Moirai \citep{woo2024unified} supports variable-dimensional inputs through any-variate attention, and Timer-XL \citep{liu2024timerxl} extends attention to autoregressive forecasting under multivariate and covariate settings.

Recent advances strengthen cross-variate modeling in time series foundation models. Chronos-2 \citep{ansari2025chronos} combines group attention with synthetic multivariate tasks, while TiRex-2 \citep{podest2026tirex2} extends masked pretraining to multivariate forecasting. Toto 2.0 \citep{khwaja2026toto2} combines temporal--variate attention with large-scale pretraining; TimesFM-3 \citep{google2026timesfm3} adopts alternating attention and parallel forecasting, and Falcon-X \citep{liu2026falconx} mediates variate interaction through latent prototypes. As these architectures accommodate diverse inputs, pretraining must expose the dependencies needed to use them. This motivates us to organize primitive compositions into forecasting episodes with different variate roles and observation masks, aligning constructed relations with observation availability across task settings.

\FloatBarrier
\section{Approach}
\label{sec:approach}

Timer-M1 aligns primitive-based data construction, model architecture, and pretraining tasks around learning temporal and relational primitives. Specifically, it composes primitives into multivariate data, organizes forecasting episodes, and learns from their permitted observations with gated two-dimensional Transformer blocks. We denote history length by \(L\), forecast horizon by \(H\), and the total number of target and covariate variates by \(C\). The model representation width is \(D\).

\begin{figure}[htbp]
\vspace{-10pt}
\centering
\includegraphics[width=\linewidth]{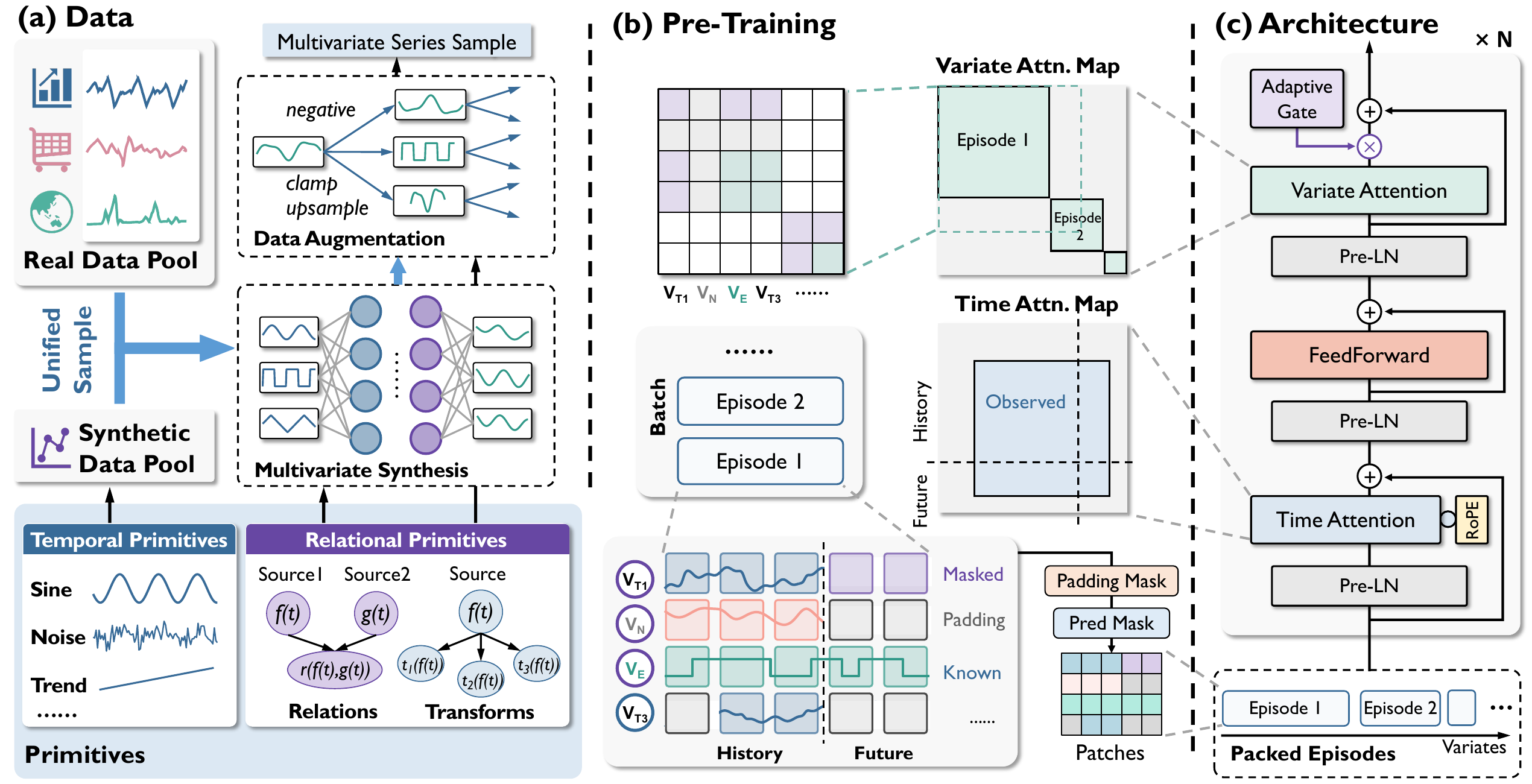}
\caption{Timer-M1 overview. (a) Temporal and relational primitives combine real and synthetic sources into multivariate samples through synthesis and augmentation. (b) Forecasting episodes specify variate roles and observation availability, with masks controlling forecasting and within-episode attention. (c) Gated two-dimensional Transformer blocks integrate temporal and cross-variate information, with learned layer-wise gates regulating cross-variate updates.}\label{fig:overview}
\end{figure}

\subsection{Data}
\label{sec:data}

Our primitive-based data pipeline brings real and synthetic series into a shared data construction process. Synthesis operators provide controlled periodic, trend, autoregressive, and other stochastic patterns. Real sources, including TimeBench \citep{liu2025sundial,liu2026timers1}, retain irregular dynamics, noise, and incomplete observations that are difficult to specify exhaustively. 

Relation operators accept either source origin or their combination, constructing signed responses, delayed effects, shared drivers, and other cross-variate dependencies. Thus, real observations serve not only as standalone training examples but also as inputs from which additional cross-variate relations can be generated. Transform operators derive varied responses from source series through sign inversion, resampling, clipping, local perturbations, and discrete-valued mappings. Applied during multivariate construction and subsequent augmentation, they diversify both cross-variate dependencies and the observed forms of primitives, with temporal transformations coordinated wherever alignment is required. Together, synthesis, relations, and transforms turn jointly sampled real and synthetic sources into diverse multivariate samples within a unified data construction pipeline. Original multivariate observations can also bypass synthesis to retain their existing dependencies.

Formally, let
\(\boldsymbol{\xi}\in\mathbb R^{\kappa\times\tau}\)
contain \(\kappa\) source series aligned over \(\tau\) consecutive time steps. Each row \(\boldsymbol{\xi}_i\) is obtained from a real-data window or a synthesis operator
\(\mathcal S_{\gamma_i}(\boldsymbol{\varepsilon}_i)\),
where \(\gamma_i\) specifies the generator parameters and
\(\boldsymbol{\varepsilon}_i\) its random inputs. A relation operator
\(\mathcal R_\phi\) produces \(\nu\) dependent response series
\(\boldsymbol{\eta}\in\mathbb R^{\nu\times\tau}\).
For branches that retain the original source channels, the resulting construction is
\begin{equation}
\begin{aligned}
\boldsymbol{\eta}
    &= \mathcal R_\phi(\boldsymbol{\xi}),\\
\boldsymbol{\Xi}
    &= \mathcal A_\psi
       \!\left([\boldsymbol{\xi};\boldsymbol{\eta}]\right).
\end{aligned}
\label{eq:data_pipeline}
\end{equation}
Here, \(\phi\) specifies the relation structure and its parameters, including mixing coefficients or temporal lags; \(\mathcal A_\psi\) denotes the selected transforms with parameters \(\psi\); and the semicolon denotes channel-wise concatenation. The output \(\boldsymbol{\Xi}\) is an aligned multivariate record from which forecasting episodes are subsequently sampled. Response-only branches omit the original source channels from the concatenation. Generation-level dimensions \(\kappa,\nu,\tau\) are distinct from the episode dimensions \(C,L,H\), which are determined after task-specific channel selection and temporal sampling.

\begin{wrapfigure}{r}{0.49\textwidth}
\vspace{-10pt}
\centering
\includegraphics[width=\linewidth]{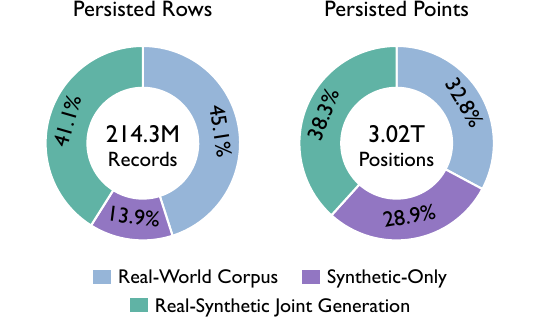}
\caption{Persisted records and scalar positions by generation route, before sampling. Real--synthetic joint generation includes corpus-derived and mixed-source generation routes.}
\label{fig:data}
\vspace{-20pt}
\end{wrapfigure}

For example, linear relations mix source channels with signed coefficients, whereas nonlinear relations apply a sampled mapping to the source vector. Reusing a source across multiple responses creates shared drivers, and lagged relations introduce delayed dependencies. The same construction applies whether the source dynamics are observed or synthesized. Representative constructions and the generation procedure are described in Appendix~\ref{sec:construction_observation}.

The stable pretraining configuration references 227 dataset entries containing 214.3 million persisted records and 3.02 trillion scalar positions (Figure~\ref{fig:data}). These unweighted storage counts precede holdouts and sampling; they are not unique observations or training exposure.

\subsection{Architecture}
\label{sec:model}

Timer-M1 uses gated two-dimensional Transformer blocks to model temporal and relational primitives. Following normalization with valid-history statistics, each variate is divided into length-\(P\) patches. A residual patch embedder maps the concatenated normalized values \(\widetilde{\mathbf x}_{c,n}\in\mathbb R^P\) and observation mask \(\mathbf m_{c,n}\in\{0,1\}^P\) to a \(D\)-dimensional token:
\begin{equation}
\mathbf h^0_{c,n}=\operatorname{PatchEmbed}\!\left([\widetilde{\mathbf x}_{c,n};\mathbf m_{c,n}]\right),
\qquad \mathbf H^0\in\mathbb R^{C\times N\times D},
\label{eq:embedding}
\end{equation}
where \(c\) indexes variates, \(n\) indexes patches, and \(N\) counts padded history and forecast patches; hidden target values and their observation masks are zeroed before embedding. Forecast tokens remain active despite their hidden values, while attention  prevents interactions across episodes.

Each layer applies temporal attention, a feed-forward network, and gated variate attention in that order. With separate pre-layer normalizations \(\operatorname{LN}\), the residual updates are
\begin{equation}
\begin{aligned}
\mathbf U^\ell &= \mathbf H^\ell + \operatorname{TimeAttention}_{\ell}\!\left(\operatorname{LN}_{t,\ell}(\mathbf H^\ell)\right),\\
\mathbf Z^\ell &= \mathbf U^\ell + \operatorname{FFN}_{\ell}\!\left(\operatorname{LN}_{f,\ell}(\mathbf U^\ell)\right),\\
\mathbf H^{\ell+1} &= \mathbf Z^\ell + \alpha_\ell\operatorname{VariateAttention}_{\ell}\!\left(\operatorname{LN}_{v,\ell}(\mathbf Z^\ell)\right),
\qquad \alpha_\ell=\sigma(a_\ell).
\end{aligned}
\label{eq:gate}
\end{equation}
Here, \(\alpha_\ell=\sigma(a_\ell)\) applies sigmoid to a learned layer-wise scalar. TimeAttention attends across patches within variates; VariateAttention attends across variates per patch. Both use multi-head scaled dot-product attention with RMS-normalized queries and keys; temporal attention uses rotary positional embeddings. The FFN comprises two linear projections with GELU. A shared quantile head maps final tokens to patch-level forecasts, restored to the original scale.

Gates adapt during training but remain fixed at inference, whereas attention weights depend on the input. Thus, the gate regulates each layer's cross-variate interaction strength rather than selecting input features. This retains standard temporal modeling while controlling the contribution of relational primitives across successive Transformer layers.

\subsection{Pre-Training}
\label{sec:episodes}

An \emph{episode} turns a generated or original multivariate record into a forecasting task by specifying history and horizon, target and covariate roles, observation availability, and future supervision. A mode describes one target and its permitted inputs; aligned modes form an episode with a shared forecast origin and attention group. Sampling different channels, windows, and roles from a record therefore exposes its underlying temporal and relational primitives under different task settings.

Write \(\mathcal E=(\mathbf x,\mathbf y,\mathbf c;\mu)\), with target history \(\mathbf x\in\mathbb R^{C_y\times L}\), target future \(\mathbf y\in\mathbb R^{C_y\times H}\), and covariates \(\mathbf c\in\mathbb R^{C_c\times(L+H)}\). Here, \(C_y\) and \(C_c\) are the numbers of target and covariate variates, with \(C=C_y+C_c\). Metadata \(\mu=(L,H,\mathbf r,\mathbf M^{\mathrm{obs}},\mathbf M^{\mathrm{loss}},g)\) specify lengths, variate roles \(\mathbf r\), observation and supervision masks, and group identity \(g\). They control packing and masking rather than supplying generating information. From permitted observations, the model predicts quantiles:

\begin{equation}
\widehat{\mathbf y}^{(q)}
=
f_\theta^{(q)}
\bigl(\mathbf x,\mathbf c;
L,H,\mathbf r,\mathbf M^{\mathrm{obs}},g\bigr),
\qquad q\in\mathcal Q.
\label{eq:episode}
\end{equation}

Here \(\theta\) denotes model parameters and \(\mathcal Q\) the set of quantile levels. Target futures are hidden. Past-only covariates are masked after the forecast origin, whereas known-future covariates remain visible where permitted by the task setting. The same variate can take different roles across episodes. Generating annotations guide the selection of derived targets and source or auxiliary variates, exposing direct, shared, and partially observed relations without revealing operator identities to the model. Additional channels may be redundant or unrelated to the prediction target. Original multivariate observations use the same interface without requiring explicit generating annotations.

Temporal sampling varies both the observed span of a primitive and the extent of its extrapolation. Episodes are packed with separate attention groups and optimized in three steps. First, define a binary supervision mask \(M^{\mathrm{loss}}_{ih}\), equal to one only at valid target-future entries, and apply horizon-decaying weights over the padded forecast length \(H_{\mathrm{pad}}\geq H\):
\begin{equation}
w_h=\frac{h^{-1}}{H_{\mathrm{pad}}^{-1}\sum_{j=1}^{H_{\mathrm{pad}}}j^{-1}},
\qquad \omega_{ih}=M^{\mathrm{loss}}_{ih}w_h.
\label{eq:loss_mask}
\end{equation}
Second, average the pointwise pinball loss over quantiles, using targets and predictions normalized by observed-history statistics:
\begin{equation}
\ell_{ih}=\frac{2}{|\mathcal Q|}\sum_{q\in\mathcal Q}
\rho_q\!\left(\widetilde y_{ih}-\widetilde{\widehat y}_{ih}^{(q)}\right),
\qquad \rho_q(v)=v\bigl(q-\mathbf1[v<0]\bigr).
\label{eq:pinball}
\end{equation}
For each target variate, tildes denote normalization of both targets and predictions using the same statistics computed from its observed history. Finally, aggregate over target variates \(i\) and forecast positions \(h\):
\begin{equation}
\mathcal L(\theta)=\frac{\sum_{i,h}\omega_{ih}\ell_{ih}}
{\sum_{i,h}\omega_{ih}+\epsilon}.
\label{eq:loss}
\end{equation}
Here \(\epsilon=10^{-8}\) stabilizes the denominator. This objective excludes unobserved supervision and emphasizes nearer forecast positions without changing the observation mask. Large-scale multivariate pretraining is followed by long-context post-training. Together, the shared objective and varied episode metadata train forecasting across different primitive compositions and observation availability. An optional GroupMix extension is described in Appendix~\ref{sec:group_mix}.

\Needspace{12\baselineskip}
\section{Experiments}
\label{sec:experiments}

\begin{wrapfigure}{r}{0.47\textwidth}
\vspace{-30pt}
\centering
\includegraphics[width=\linewidth]{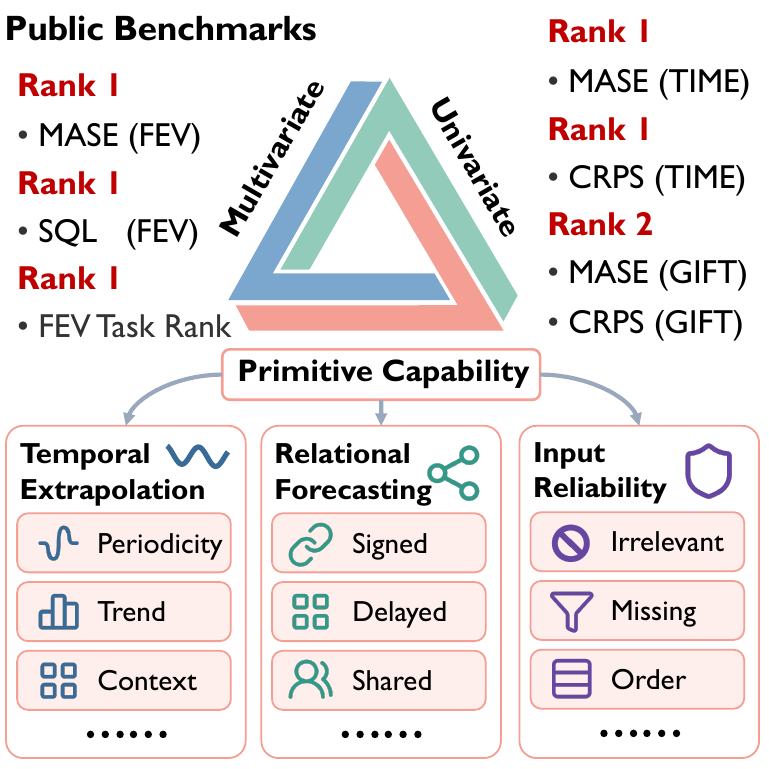}
\caption{Evaluation overview. Public benchmarks assess forecasting across settings, while primitive evaluations examine temporal extrapolation, relational forecasting, and input reliability. Rankings are among compared methods.}\label{fig:evaluation}
\vspace{-45pt}
\end{wrapfigure}

We evaluate Timer-M1 through public benchmarks, primitive evaluation, and ablation studies. These examine forecasting accuracy, temporal extrapolation, relational forecasting, and input reliability, alongside the contributions of adaptive gates and data construction. Together, they connect performance with learning reusable primitives through coordinated data construction, episode-based pretraining, and model design.

\Needspace{9\baselineskip}
\subsection{Public Benchmarks}
\label{sec:public}

We evaluate 100 FEV \citep{shchur2025fev}, 97 GIFT-Eval \citep{aksu2024gifteval}, and 98 TIME \citep{qiao2026time} configurations, reporting point and probabilistic scores over matched sets. Public baseline submissions are compared with a shared Timer-M1 checkpoint. FEV distinguishes five task types, while TIME and predominantly univariate GIFT-Eval complement coverage across temporal settings. Details are provided  in Appendix~\ref{sec:benchmark_details}.

\begin{figure}[htbp]
\centering
\includegraphics[width=\linewidth]{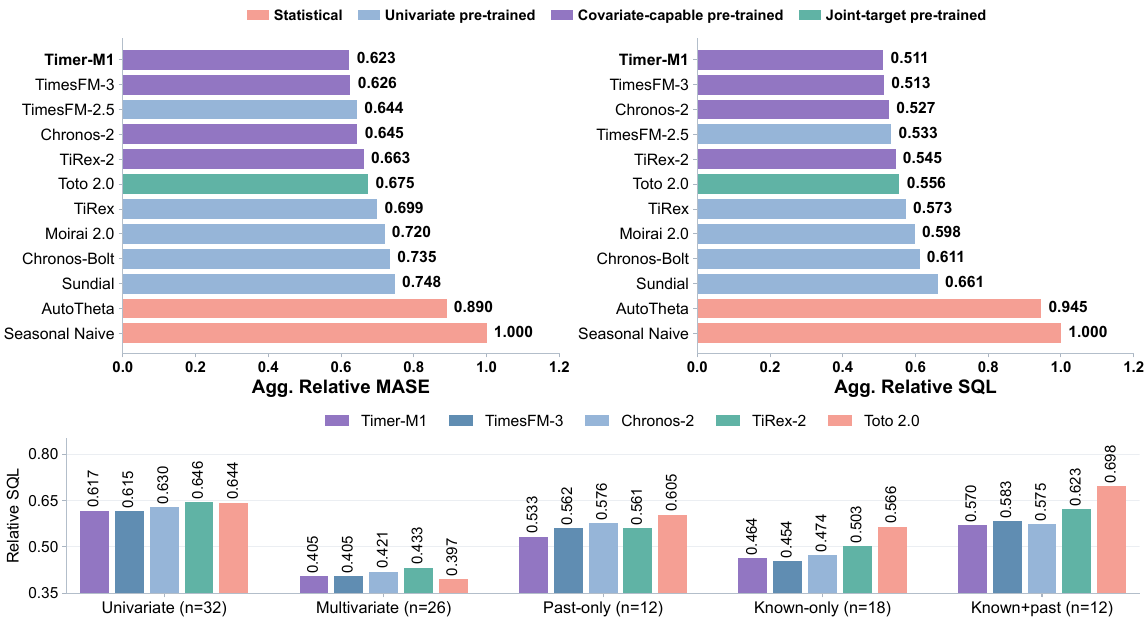}
\caption{Forecasting results on FEV across 100 matched configurations. Top: aggregate MASE and SQL for 12 methods, normalized by Seasonal Naive. Bottom: SQL across five task types with different variate roles and observation availability. Timer-M1 achieves the lowest aggregate errors. Colors denote forecasting interfaces above and individual models below.}
\label{fig:fev}
\vspace{-10pt}
\end{figure}

\paragraph{FEV.}
Timer-M1 achieves MASE skill 0.3771 and SQL skill 0.4887 over the 100 matched configurations, ranking first on both aggregate metrics among the compared methods. Its relative errors are lower than those of Chronos-2 by 3.43\% and 3.02\%, and lower than those of TimesFM-3 by 0.47\% and 0.42\%, respectively. Figure~\ref{fig:fev} further separates five mutually exclusive task types: univariate, multivariate, past-only covariates, known-future covariates, and their combination. Across these groups, Timer-M1 achieves the best unweighted average rank for both MASE (1.80) and SQL (1.60), placing among the top two models in every group under both metrics. This consistency complements its aggregate accuracy, demonstrating broad task coverage within a compact 121M-parameter backbone. A shared frozen checkpoint and inference recipe accommodate different variate roles and observation conditions without task-specific parameter updates, connecting temporal and relational primitives with flexible episode construction.

\begin{figure}[htbp]
\centering
\includegraphics[width=\linewidth]{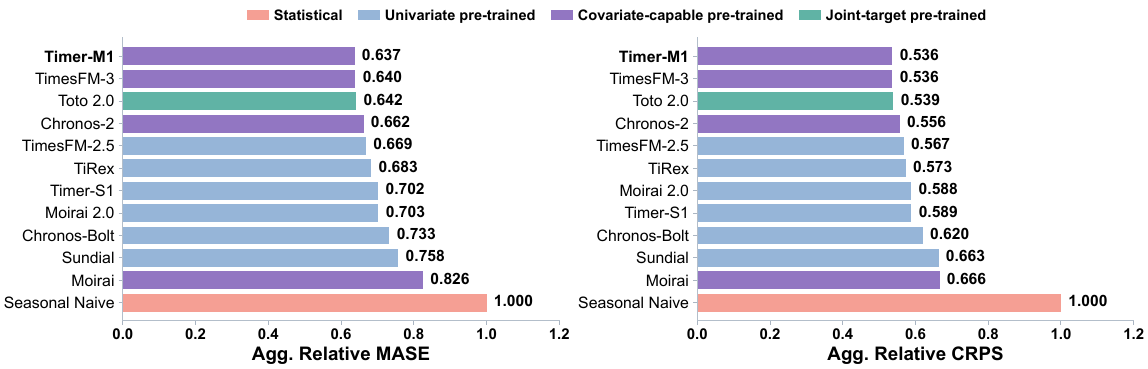}
\caption{Comparison on TIME across 98 matched configurations. Timer-M1 achieves state-of-the-art aggregate performance, attaining the lowest normalized MASE and CRPS among the compared methods across diverse forecasting task settings. Colors distinguish forecasting interfaces.}
\vspace{-10pt}
\label{fig:time}
\end{figure}

\paragraph{TIME.}
Timer-M1 obtains normalized MASE 0.6372 and CRPS 0.5363 over 98 configurations (Figure~\ref{fig:time}), reducing errors relative to Chronos-2 by 3.75\% and 3.60\%, respectively. These improvements extend across both point and probabilistic aggregate scores, suggesting benefits beyond point estimation alone, complementing the results on FEV across diverse forecasting task settings. For comparison with TimesFM-3, a repeat under the same evaluation conditions yields 0.63977 and 0.53635, corresponding to reductions of 0.40\% and less than 0.01\%. We interpret this narrow margin as competitive performance with TimesFM-3 rather than statistically established superiority, while observing larger improvements over Chronos-2 on both reported metrics.

\begin{figure}[htbp]
\centering
\includegraphics[width=\linewidth]{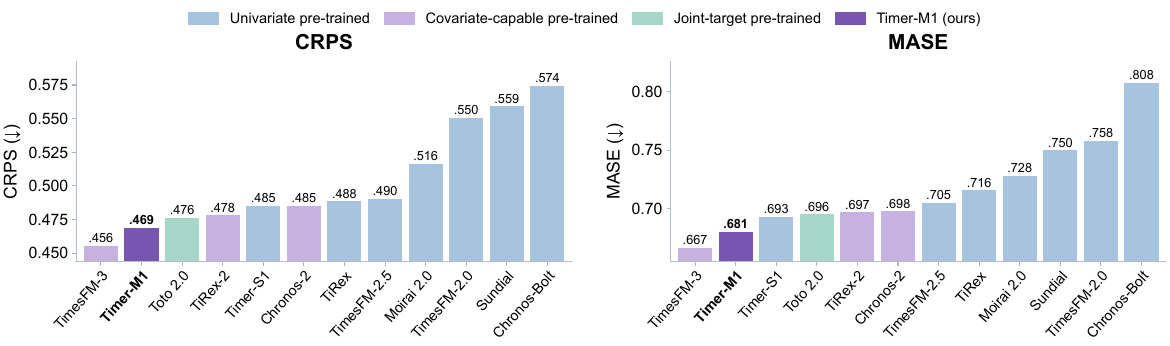}
\caption{Point and probabilistic forecasting performance on GIFT-Eval across 97 matched configurations. Timer-M1 ranks second among 12 compared pretrained models on both MASE and CRPS, demonstrating competitive generalization across diverse domains and forecast horizons.}
\label{fig:gift}
\end{figure}

\paragraph{GIFT-Eval.}
Across 97 configurations, Timer-M1 obtains normalized MASE 0.6806 and CRPS 0.4688 (Figure~\ref{fig:gift}), reducing errors relative to Chronos-2 by 2.47\% and 3.43\%, respectively. Timer-M1 also improves both metrics over TiRex-2 and Toto 2.0 (2.5B), ranking second among the compared methods, while TimesFM-3 retains the lowest aggregate errors. The benchmark spans diverse sampling frequencies and forecast lengths, testing whether a shared pretrained model remains effective across heterogeneous temporal resolutions and task settings. These results on predominantly univariate GIFT-Eval complement FEV and TIME, showing that Timer-M1's improvements extend beyond multivariate tasks to both point and probabilistic forecasting of individual series.

\subsection{Primitive Evaluation}
\label{sec:primitive_eval}

\begin{figure}[t]
\centering
\includegraphics[width=\linewidth]{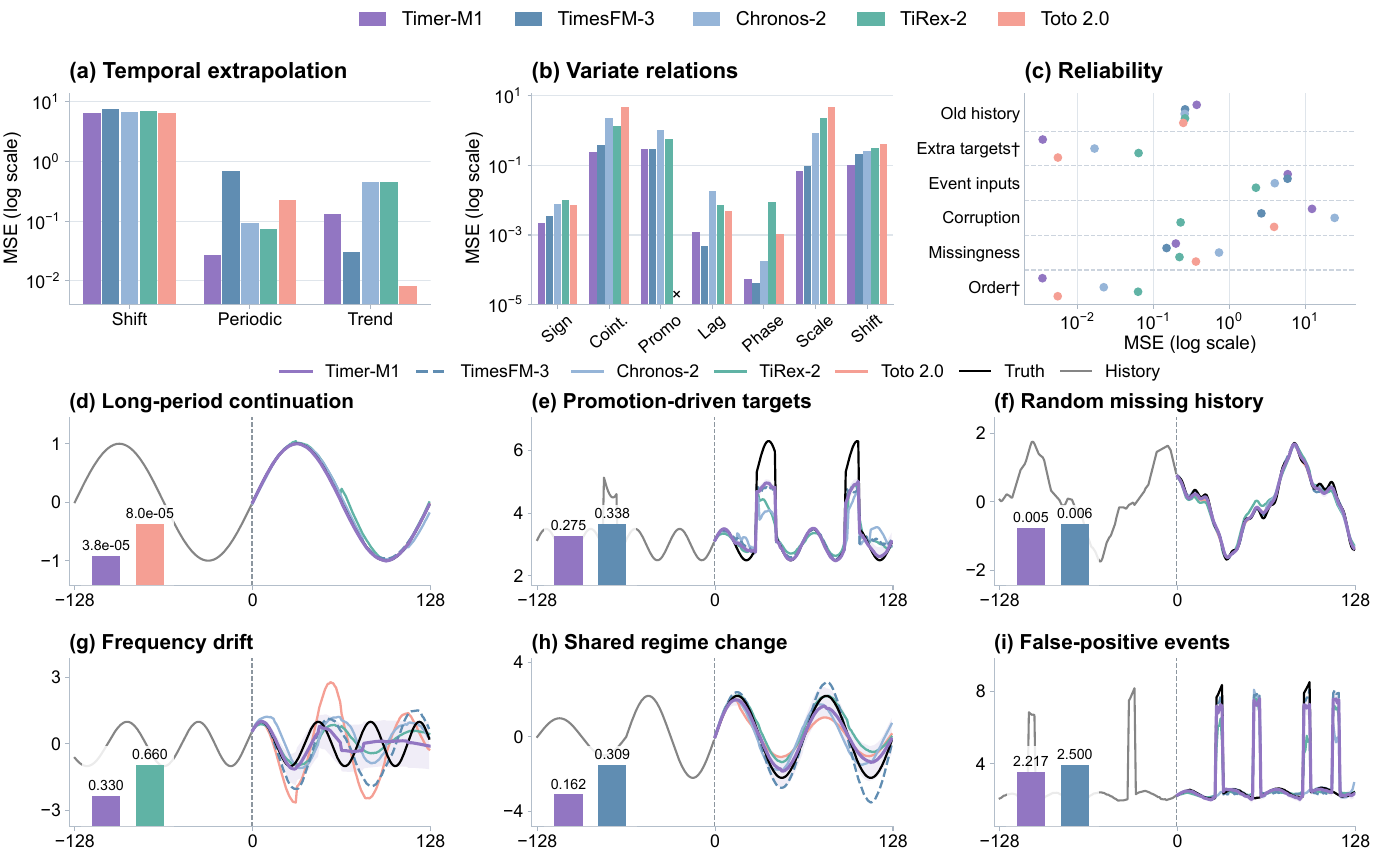}
\caption{Primitive evaluation. Top: predefined temporal groups, seven relational mechanisms, and six reliability categories, measured by MSE (lower is better). Unsupported interfaces are omitted; detailed results for the displayed groups appear in Appendix~\ref{sec:diagnostic_details}. Bottom: six strength cases spanning distinct mechanisms, not an estimate of their frequency. Shading denotes M1's marginal 10--90\% quantiles. $\dagger$ marks unavailable squared-error archives for some baseline.}
\label{fig:primitives}
\end{figure}

Primitives encompass recurring temporal and relational patterns that support forecasting across task settings. To examine how Timer-M1 learns and applies these primitives, we conduct \emph{Primitive Evaluation}, assessing temporal extrapolation, relational forecasting, and input reliability through controlled tasks. Figure~\ref{fig:primitives} presents aggregate results over complete task groups alongside illustrative cases, connecting forecasting performance with responses to systematic changes in temporal patterns, variate relations, and observation availability, thereby examining primitive reuse under changing conditions. Details are provided in Appendices~\ref{sec:primitive_protocol}, \ref{sec:diagnostic_details}, and~\ref{sec:showcases}.

\paragraph{Temporal extrapolation.}
Timer-M1 shows strong periodic extrapolation, achieving NMAE $0.02432$ across 13 periodic cases. Long-period continuation and frequency drift illustrate its ability to follow recurring patterns beyond the observed context. These results are consistent with learning reusable temporal primitives across diverse compositions, linking accurate periodic continuation with the ability to track evolving temporal patterns rather than repeat observed sequences.

\paragraph{Relational forecasting.}
Timer-M1 achieves NMAE $0.02160$ across six joint-target tasks, demonstrating strong cross-variate forecasting. Promotion-driven targets and shared regime changes illustrate how relational primitives connect observations to target dynamics. These findings support learning reusable relational primitives through data construction and episode-based pretraining, which exposes dependencies under different variate roles and observation availability.

\paragraph{Input reliability.}
Timer-M1 reduces target-missingness NMAE by $32.5\%$ relative to Chronos-2, with selected cases illustrating resilience to incomplete histories and misleading inputs. These findings support composing primitives under varied observation conditions to broaden coverage of challenging forecasting scenarios. Episodes with observation masks and redundant or unrelated variates complement this construction by exposing primitives through imperfect inputs.

\subsection{Ablation Study}
\label{sec:ablation}
\vspace{11pt}

\begin{figure}[htbp]
\centering
\makeatletter
\long\def\@makecaption#1#2{\vskip\abovecaptionskip\noindent#1: #2\par}
\makeatother
\begin{minipage}[t]{0.46\linewidth}
\vspace{0pt}
\begin{minipage}[t][104pt][t]{\linewidth}
\vspace{0pt}
\centering
\includegraphics[width=\linewidth]{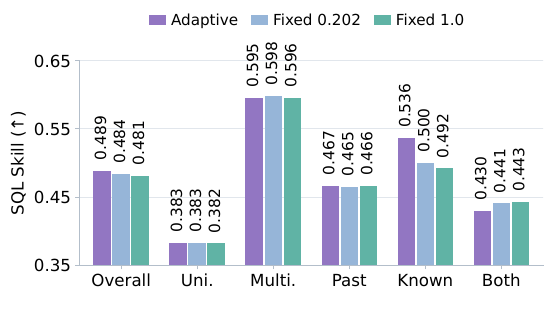}
\end{minipage}
\caption{Gate ablation on overall FEV and five task types. Improvement supports layer-wise allocation of cross-variate interaction.}
\label{fig:gate_ablation}
\end{minipage}\hfill
\begin{minipage}[t]{0.52\linewidth}
\vspace{0pt}
\begin{minipage}[t][104pt][t]{\linewidth}
\vspace{0pt}
\centering
\fontsize{8.5}{10}\selectfont
\renewcommand{\arraystretch}{1.12}
\setlength{\tabcolsep}{2pt}
\begin{tabular*}{\linewidth}{@{\extracolsep{\fill}}lcccc@{}}
\toprule
Metric & \makecell{Real\\Only} & \makecell{Synthetic\\Only} & \makecell{No-Joint} & \makecell{Full\\Recipe} \\
\midrule
FEV Skill Score  $\uparrow$ & 0.4431 & 0.4710 & \secondres{0.4725} & \textcolor{red}{\textbf{0.4887}} \\
\addlinespace[2pt]
GIFT MASE $\downarrow$ & 0.6998 & 0.6884 & \secondres{0.6819} & \textcolor{red}{\textbf{0.6806}} \\
GIFT CRPS $\downarrow$ & 0.4850 & 0.4741 & \secondres{0.4722} & \textcolor{red}{\textbf{0.4688}} \\
\addlinespace[2pt]
TIME MASE $\downarrow$ & 0.6579 & 0.6438 & \secondres{0.6379} & \textcolor{red}{\textbf{0.6372}} \\
TIME CRPS $\downarrow$ & 0.5526 & 0.5456 & \textcolor{red}{\textbf{0.5359}} & \secondres{0.5363} \\
\bottomrule
\end{tabular*}
\end{minipage}
\makeatletter\def\@captype{table}\makeatother
\caption{Data recipe ablation. Red bold and blue underline mark the best and second-best scores in each row. Joint generation benefits most aggregates.}\label{tab:ablation}
\end{minipage}
\end{figure}

We conduct ablation studies to examine how adaptive cross-variate interaction and primitive-based data construction contribute to forecasting. Gate ablations compare learned gates with fixed alternatives, testing the benefit of allocating interaction strength across layers. Data ablations compare four recipes to assess the contribution of joint real--synthetic generation beyond source diversity. All variants use the same evaluation inputs and inference recipe. Figure~\ref{fig:gate_ablation} reports overall and task-specific FEV results, while Table~\ref{tab:ablation} summarizes data ablations across benchmarks.

\vspace{1pt}

\paragraph{Adaptive gates.}
We compare learned layer-wise coefficients $\alpha_\ell$ with two uniform alternatives. \emph{Fixed 0.202} uses the mean gate value from the stable checkpoint, preserving a representative interaction strength while removing layer-wise variation; \emph{Fixed 1.0} applies unattenuated cross-variate updates. Adaptive gates achieve FEV SQL skill $0.4887$, versus $0.4836$ and $0.4814$, improving three of five task types. The learned $\alpha_\ell$ generally increase with depth, suggesting stronger cross-variate integration in deeper layers. Together, this depth-dependent pattern and the advantage over mean-matched scaling support allocating interaction strength across layers, allowing relational primitives to be integrated progressively with temporal representations.

\vspace{11pt}
\paragraph{Data construction.}
We compare \emph{Real Only}, \emph{Synthetic Only}, \emph{No-Joint}, and the \emph{Full Recipe}. The first two restrict generation to their respective source pools. No-Joint includes both sources but excludes joint real--synthetic generation; the full recipe additionally composes them through shared relation operators. The full recipe improves FEV skill and both GIFT-Eval metrics over all three alternatives and achieves the lowest TIME MASE. Its FEV skill exceeds No-Joint by $0.0162$, suggesting benefits beyond source diversity alone. By combining observed dynamics with controlled relational primitives, joint generation broadens multivariate compositions, supporting a unified data construction pipeline rather than independent real and synthetic training collections.

\FloatBarrier
\vspace{11pt}
\section{Conclusion}

We present Timer-M1, a multivariate time series foundation model that learns with temporal and relational primitives. Its primitive-based data pipeline combines observed dynamics with controlled generation to construct diverse multivariate samples. Episode-based pretraining exposes these primitives under varied variate roles and observation availability, while gated two-dimensional Transformer blocks integrate temporal and cross-variate information. Evaluations on three public benchmarks and controlled primitive tasks demonstrate competitive forecasting performance and strengths in temporal extrapolation and relational forecasting. These results support aligning data construction, forecasting episodes, and model architecture around shared primitives. Future work will expand real-data coverage and primitive compositions to better represent irregular dynamics and partially observed dependencies in complex real-world systems. On the architectural side, exploring input-dependent cross-variate interaction and more efficient long-context modeling may further improve how foundation models learn and apply primitives across domains and task settings.

\clearpage

\subsection*{AI use statement}

Generative AI tools were used only for translation, drafting and revising portions of the manuscript, language polishing, literature search, reference formatting, and figure preparation. They were not used for developing theoretical models or conceptual frameworks, designing research methodology or experiments, writing proofs, implementing experimental methods, generating or processing datasets, conducting data analysis, or interpreting results, and no substantive AI-generated suggestions regarding mathematical assumptions, derivations, proofs, or research hypotheses were adopted. All AI-assisted work was reviewed by the authors, who consulted original references, checked derivations and experimental code, and verified reported results against experimental records; we take full responsibility for the final content of this work.

\bibliography{references}
\bibliographystyle{timer_m1_arxiv}
\clearpage
\appendix
\section{Dataset Statistics}
\label{sec:training_details}

This section describes the temporal and relational primitives used by the data construction pipeline. The discussion complements Section~\ref{sec:data} by connecting operator families to the temporal evolution and cross-variate dependencies that define a forecasting sample. Real observations retain empirical dynamics, while generated series make selected patterns and dependencies controllable. Both enter the same multivariate construction process, allowing empirical dynamics and controlled primitive compositions to contribute to a shared source of forecasting episodes.

\subsection{Temporal and Relational Primitives}
Temporal primitives describe evolution within a series. Periodic and trend components may appear alone or together with stochastic variation; autoregressive dependence provides a further source of temporal continuity. Relational primitives describe how multiple variates evolve together. Source reuse introduces shared drivers, signed mixing produces positive or negative responses, and temporal offsets create delayed effects. These families are complementary: temporal primitives determine the evolution of individual sources, while relational primitives determine how their information is shared across variates. For example, a periodic source can produce responses with different signs, scales, or delays without changing its underlying temporal pattern. The resulting dependencies are specified during construction rather than inferred as causal mechanisms from real observations.
\begin{table}[htbp]
\centering
\caption{Operator families in primitive-based data construction. The taxonomy follows the temporal and relational primitives introduced in the main text. Examples describe mechanisms rather than a fixed generation recipe.}
\label{tab:operators}
\small
\renewcommand{\arraystretch}{1.18}
\setlength{\tabcolsep}{3pt}
\begin{tabular*}{\textwidth}{@{\extracolsep{\fill}}llp{0.51\textwidth}@{}}
\toprule
Family & Primitive & Representative construction \\
\midrule
\multirow{4}{*}{Temporal}
& Periodicity & Oscillatory components and their superposition \\
& Trend & Gradual growth, decline, and shared drift \\
& Stochastic dependence & Noise and autoregressive dynamics \\
& Regime changes & Piecewise changes in level or evolution \\
\midrule
\multirow{6}{*}{Relational}
& Signed responses & Positive and negative source mixing \\
& Delayed responses & Temporal offsets between source and response \\
& Shared drivers & Multiple responses derived from common sources \\
& Cointegration & Shared nonstationary components \\
& Event effects & Responses to discrete-valued inputs \\
& Transforms & Inversion, clipping, resampling, and discrete mappings \\
\bottomrule
\end{tabular*}
\end{table}

\subsection{Construction and Observation}
\label{sec:construction_observation}

Following Equation~\ref{eq:data_pipeline}, source series are selected from real observations or generated temporal primitives, combined through relation operators, and transformed into aligned multivariate records. For source array \(\boldsymbol{\xi}\in\mathbb R^{\kappa\times\tau}\), a schematic linear response is
\begin{equation}
\eta_i(t)=\sum_{j\in\mathcal J_i}w_{ij}\xi_j(t-\delta_{ij}),
\qquad i=1,\ldots,\nu,
\label{eq:appendix_composition}
\end{equation}
where \(\mathcal J_i\subseteq\{1,\ldots,\kappa\}\) selects the sources contributing to response \(i\), \(w_{ij}\) controls their signed contribution, and \(\delta_{ij}\) denotes a temporal lag. Nonlinear relation operators replace the linear combination with a mapping of the selected sources. Reusing a source across responses introduces shared variation. The transform operator \(\mathcal A_\psi\) then changes the observed forms through operations such as rescaling or clipping. For branches retaining source channels, the resulting record is \(\boldsymbol{\Xi}=\mathcal A_\psi([\boldsymbol{\xi};\boldsymbol{\eta}])\); response-only branches omit \(\boldsymbol{\xi}\). Transformations that modify time coordinates are coordinated where alignment must be preserved. Original multivariate observations can also enter directly, retaining their existing dependencies.

A multivariate record specifies aligned values, whereas a forecasting episode additionally specifies their roles and availability. After channel selection and temporal sampling, let \(\mathbf X\in\mathbb R^{C\times(L+H)}\) contain the episode's target and covariate values. Following the main text, \(\mathbf M^{\mathrm{obs}}\in\{0,1\}^{C\times(L+H)}\) denotes the observation mask, while \(\mathbf M^{\mathrm{loss}}\in\{0,1\}^{C_y\times H}\) selects supervised target-future entries. The masked values are defined entrywise as
\begin{equation}
X^{\mathrm{obs}}_{c,t}=
\begin{cases}
X_{c,t}, & M^{\mathrm{obs}}_{c,t}=1,\\
0, & M^{\mathrm{obs}}_{c,t}=0,
\end{cases}
\qquad
M^{\mathrm{loss}}_{ih}=
\begin{cases}
1, & \text{if target }i\text{ at horizon }h\text{ is valid},\\
0, & \text{otherwise}.
\end{cases}
\label{eq:appendix_observation}
\end{equation}
Here, \(c\) indexes all episode variates, \(t\) indexes history and forecast positions, \(i\) indexes target variates, and \(h\) indexes forecast positions. Normalization uses observed-history statistics, and hidden entries remain zero in the values supplied to the patch embedder. The observation mask accompanies these values so that hidden entries remain distinguishable from observed zeros. Target futures are withheld from the input but may contribute to supervision; known-future covariates remain visible where permitted. Different role assignments therefore expose the same primitives through different forecasting conditions without changing the underlying record.

The record and scalar-position counts in Figure~\ref{fig:data} describe stored data before episode sampling. Records, channel-wise series, and scalar positions are distinct counting units. If record \(r\) contains \(C_r\) channels of length \(T_r\), their respective counts are
\begin{equation}
N_{\mathrm{records}}=\sum_r 1,
\qquad
N_{\mathrm{series}}=\sum_r C_r,
\qquad
N_{\mathrm{positions}}=\sum_r C_rT_r.
\label{eq:data_counts}
\end{equation}
These quantities characterize stored data volume rather than independent forecasting tasks or consumed training tokens. Episode sampling can reuse a record under different horizons and observation roles.

\FloatBarrier
\section{Implementation Details}
\label{sec:architecture_details}
\subsection{Model Configuration}
Timer-M1 uses the same gated two-dimensional Transformer backbone across task settings. Table~\ref{tab:model_configuration} summarizes its dimensions. Temporal attention operates along patches within each variate, and variate attention operates across variates at each patch position. The residual order is temporal attention, feed-forward transformation, and gated variate attention. The prediction head outputs marginal quantiles, with the median used for point forecasts.
\begin{table}[htbp]
\centering
\caption{Model configuration. The gate contributes one learned scalar per layer; the total parameter count includes the embedding and prediction head.}
\label{tab:model_configuration}
\small
\renewcommand{\arraystretch}{1.18}
\setlength{\tabcolsep}{3pt}
\begin{tabular*}{\textwidth}{@{\extracolsep{\fill}}lc@{}}
\toprule
Component & Configuration \\
\midrule
Parameters & 120,998,412 (approximately 121M) \\
Transformer layers & 12 \\
Representation / feed-forward width & 768 / 3072 \\
Attention heads / head dimension & 12 / 64 \\
Patch length & 16 observations \\
Prediction quantiles & $0.1,0.2,\ldots,0.9$ \\
Temporal position encoding & Rotary positional embeddings \\
Gate parameterization & $\alpha_\ell=\sigma(a_\ell)$; one scalar per layer \\
Gate initialization & $a_\ell=-4.6$, giving $\alpha_\ell\approx0.01$ \\
\bottomrule
\end{tabular*}
\end{table}

For $C$ variates, $N$ patches, and width $D$, factorized attention requires $\mathcal O(CN^2D+NC^2D)$ operations per layer, rather than $\mathcal O(C^2N^2D)$ for attention over all flattened tokens. Linear projections and feed-forward transformations additionally require $\mathcal O(CND^2)$ operations. Gating adds a scalar multiplication to each variate-attention residual update. The same patch embedder and prediction head are shared across variates, while episode boundaries determine which cross-variate interactions are permitted. This separates the number and roles of input variates from the learned representation width.

\subsection{Layer-wise Gates}
\begin{figure}[htbp]
\centering
\includegraphics[width=0.94\linewidth]{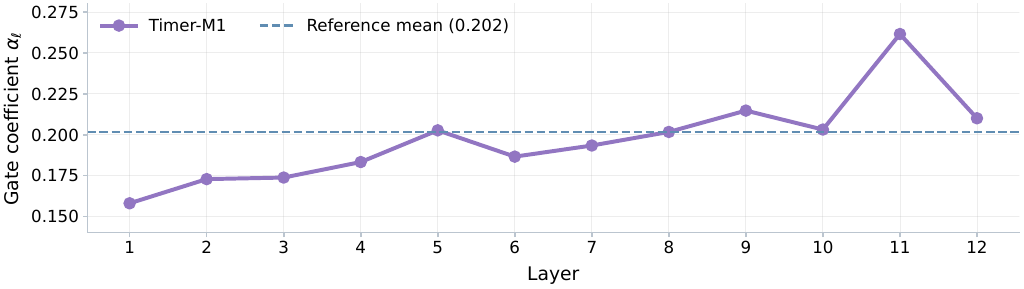}
\caption{Layer-wise allocation of cross-variate interaction. The learned coefficients generally rise with depth, while local deviations allow different layers to assign different weights to variate-attention updates. The dashed line marks the earlier stable-model mean used by the fixed-$0.202$ control, not the mean of the displayed curve. Together with the gate ablation, the profile motivates adaptive allocation rather than a uniform residual scale.}
\label{fig:gate_depth}
\end{figure}

Figure~\ref{fig:gate_depth} shows the learned coefficients $\alpha_\ell=\sigma(a_\ell)$ of the evaluated model. Each coefficient scales the variate-attention residual update after temporal attention and the feed-forward transformation. It is shared across tokens in a layer and remains fixed during inference; it is not a sample-dependent attention score. The coefficients generally increase with depth, with larger updates in later layers and nonmonotonic local variation. This pattern is consistent with first forming temporal representations and subsequently giving greater weight to cross-variate integration, although coefficient magnitudes alone do not measure the information carried by either branch.

The fixed value $0.202$ is the rounded mean of twelve gate coefficients from an earlier stable model. It preserves a representative interaction scale while removing layer-specific allocation; the other fixed control, $1.0$, uses the unscaled variate-attention update. As shown in Table~\ref{tab:gate_benchmarks}, the learned gates improve overall FEV skill.

\subsection{Episodes and Packing}
Episodes share a forecast origin and specify target, past-only covariate, and known-future covariate roles. Target futures are hidden; available future covariates remain visible. Observation masks determine which values enter the model, while supervision masks select valid target-future entries for the loss. Padding is excluded from valid observations without disabling the forecast tokens that must produce predictions.

We formulate episode batching as a capacity-constrained bin-packing problem and use a greedy packing heuristic. Episodes remain intact, and distinct attention groups prevent information exchange between packed episodes. Packing changes how episodes share a batch, not their permitted observations or prediction targets. The packing objective improves batch utilization while preserving episode semantics; the greedy procedure does not require an exact solution of the bin-packing problem.

\paragraph{GroupMix.}
\label{sec:group_mix}
GroupMix extends episode construction by changing cross-variate visibility during training. Starting from separate episode groups, a selected group is associated with additional variates from other groups in the packed batch. In the group-level form, the selected groups share an attention group, permitting interaction among their visible variates. A finer-grained form introduces only a subset of external variates. Sequence values, variate roles, and supervision targets are retained: the operation modifies the attention grouping rather than interpolating series or mixing forecast labels.

Observation and supervision masks continue to enforce the original information constraints. In particular, hidden target futures are not made visible by a group reassignment, and padded entries do not become observations. Unrelated external variates can consequently act as distractors, providing a way to vary the context in which temporal and relational primitives must be recognized. GroupMix is a training-time extension; ordinary episode grouping is used at inference. It is not enabled for the results reported here, so its potential regularization effect is distinct from the measured gate and data-construction ablations.

\subsection{Public Benchmark Protocols}
\label{sec:benchmark_details}
We follow the official task definitions and scoring conventions of FEV, TIME, and GIFT-Eval~\citep{shchur2025fev,qiao2026time,aksu2024gifteval}. Forecast origins, prediction horizons, and permitted observations are determined by the benchmark tasks. FEV distinguishes historical covariates from future-known covariates; only inputs available at the forecast origin are supplied. Scores use the complete matched configuration sets reported in the main text. Public reference scores are fixed comparison snapshots, not a continuously updated ranking.

The official online evaluation documentation was checked on September 26, 2026.\footnote{\href{https://github.com/autogluon/fev}{FEV}, \href{https://github.com/zqiao11/TIME}{TIME}, and \href{https://github.com/SalesforceAIResearch/gift-eval}{GIFT-Eval}.} Compliance with task definitions and metrics is distinct from training-data eligibility. No task-specific parameter updates are performed during evaluation. Training-data exposure is not matched across the compared methods.

For model $m$ and configuration $j$, let $E_{m,j}$ denote the benchmark error and $E_{\mathrm{SN},j}$ the Seasonal Naive error. For a configuration group $G$, the relative aggregate is
\begin{equation}
R_{m,G}=\exp\!\left(\frac{1}{|G|}\sum_{j\in G}\log\frac{E_{m,j}}{E_{\mathrm{SN},j}}\right).
\label{eq:benchmark_metric}
\end{equation}
FEV clips each relative error to $[0.01,100]$ before aggregation and reports skill $1-R_{m,G}$. We report MASE and SQL skill separately. TIME and GIFT-Eval report normalized MASE and quantile-based CRPS. Higher skill and lower normalized error indicate better forecasts. Missing matched results remain unavailable rather than being assigned zero error.

\subsection{Primitive Evaluation Protocol}
\label{sec:primitive_protocol}
The controlled evaluation separates temporal extrapolation, relational forecasting, and input reliability. Each test fixes a generated trajectory and forecast origin, then specifies the observations available to each model. All compared models receive the same target values and permitted covariates for a given setting. Future targets are used only for scoring. The displayed groups are defined by their mechanism, not by the model attaining the smallest error, and retain all scoring records within each selected group.

\paragraph{Temporal extrapolation.}
The elementary-function tests use $L=256$ historical observations and $H=128$ forecast steps. Periodicity covers sinusoidal and nonsinusoidal waves, multiple frequencies, amplitude modulation, frequency drift, and narrow pulses. Trend tests include constant, increasing, decreasing, nonlinear, and piecewise trajectories. The shift group includes a level change visible in history and changes in level, slope, amplitude, frequency, or phase introduced at the forecast boundary. The latter have no advance indication in the supplied history and are interpreted as stress tests, not as evidence that a model can anticipate an unobserved intervention. Table~\ref{tab:primitive_protocol} connects these mechanisms to the result groups.

\paragraph{Relational forecasting.}
Structured tasks use the same $L=256$, $H=128$ window and construct multiple responses from shared sources. Signed, delayed, phase-shifted, scale-varying, cointegrated, and shared-regime relations probe different ways of transferring information across variates. The joint-target settings hide all target futures; promotion-driven settings additionally supply the event covariate over its permitted interval. Each target is scored separately before group aggregation. These cross-model comparisons assess forecasting with the specified inputs; they are not, by themselves, estimates of the gain from joint rather than independent prediction.

\paragraph{Input reliability.}
These tests use $L=512$ and $H=128$ and hold the clean future fixed while changing the available history or auxiliary inputs. Missing-history settings include random removal at $10\%$, $30\%$, and $50\%$, internal gaps of 32 or 96 steps, and missing suffixes of 16 or 64 steps. Internal gaps use common linear interpolation; missing suffixes use forward filling, without consulting future targets. Corruption tests perturb the observed history; irrelevant-input tests introduce unrelated or misleading context. Event tests alter the supplied event indicators while retaining the same target trajectory. Variate-order tests permute the same input channels and restore predictions to the original target order before scoring. Thus, missingness results concern forecasting after the specified common preprocessing, rather than a separate imputation task.

\begin{table}[!ht]
\centering
\caption{Displayed primitive evaluation groups and their controlled conditions. The counts of scoring records and model-specific results appear in Tables~\ref{tab:primitive_temporal}--\ref{tab:primitive_reliability}. All groups report MSE; NMAE additionally supports the comparisons discussed in the text.}
\label{tab:primitive_protocol}
\small
\renewcommand{\arraystretch}{1.16}
\setlength{\tabcolsep}{3pt}
\begin{tabular*}{\textwidth}{@{\extracolsep{\fill}}lp{0.23\textwidth}p{0.51\textwidth}@{}}
\toprule
Capability & Group & Controlled pattern or observation condition \\
\midrule
Temporal & Periodic & Period, waveform, modulation, and gradual frequency change \\
 & Trend & Direction, growth shape, and observed slope change \\
 & Shift & Observed level change and unannounced boundary changes \\
\midrule
Relational & Sign; scale & Shared dynamics with opposite signs or different scales \\
 & Lag; phase & Delayed responses and phase-shifted periodic targets \\
 & Cointegration & Multiple responses sharing a nonstationary source \\
 & Promotion; shift & Event-driven responses and shared regime changes \\
\midrule
Reliability & Irrelevant context & Uninformative or misleading portions of historical context \\
 & Irrelevant targets & Additional unrelated variates in the joint target set \\
 & Missing event inputs & Incomplete or incorrect future event information \\
 & Target corruption & Noise and outlying values in the observed target history \\
 & Target missingness & Random gaps, internal blocks, and missing recent history \\
 & Variate order & Permutation of input channels with target identities restored \\
\bottomrule
\end{tabular*}
\end{table}

\paragraph{Metrics and aggregation.}
For scoring record $r$, let $\Omega_r$ contain its evaluated forecast positions and let $\hat y_{r,i}$ and $y_{r,i}$ denote the point forecast and ground truth. We compute
\begin{equation}
\operatorname{MSE}_r=\frac{1}{|\Omega_r|}\sum_{i\in\Omega_r}(\hat y_{r,i}-y_{r,i})^2,
\qquad
\operatorname{NMAE}_r=\frac{\sum_{i\in\Omega_r}|\hat y_{r,i}-y_{r,i}|}{|\Omega_r|s_r},
\label{eq:primitive_metrics}
\end{equation}
where $s_r$ is a fixed reference scale shared by the models for that record. The displayed short-window temporal and relational groups use the range of the complete generated trajectory, including history and future. For a nearly constant trajectory, the denominator is the larger of its mean absolute value and one. Reliability tests use the range of the clean historical reference with a positive floor of $10^{-6}$. These quantities normalize evaluation only and are not provided to the forecasting model. Group scores are arithmetic means over their scoring records, with one record per evaluated target and condition. The count $n$ is therefore not necessarily the number of independent generated trajectories. Rankings are computed before rounding; red bold and blue underline indicate the lowest and second-lowest distinct errors, respectively.

MSE retains the effect of absolute error scale, whereas NMAE facilitates comparison between trajectories of different amplitudes. The two metrics are not interchangeable, and no cross-capability composite score is inferred from the selected groups. Unsupported input interfaces and unavailable metrics are left unranked rather than assigned an artificial penalty. Where quantiles are available, showcase shading spans the 10th to 90th percentiles; these are marginal prediction intervals, not confidence intervals for the estimated mean. Examples illustrate individual mechanisms and do not determine the numerical aggregates.

\clearpage
\section{Full Results}
\subsection{Public Benchmarks}
Table~\ref{tab:public_scores} provides the numerical aggregates behind the public-benchmark comparisons. Scores retain the original benchmark direction and normalization. The comparisons cover 100 FEV, 98 TIME, and 97 GIFT-Eval configurations. Training-data eligibility is described in Appendix~\ref{sec:benchmark_details}; numerical ranks refer to the displayed methods rather than a claim of official zero-shot leaderboard placement.
\begin{table}[htbp]
\centering
\caption{Aggregate benchmark results. FEV reports skill ($\uparrow$); TIME and GIFT-Eval report normalized errors ($\downarrow$). Red bold and blue underline mark the best and second-best displayed scores per column. Dashes indicate unavailable matched results. Training exposure differs across methods (Appendix~\ref{sec:benchmark_details}).}
\label{tab:public_scores}
\small
\renewcommand{\arraystretch}{1.18}
\setlength{\tabcolsep}{3pt}
\begin{tabular*}{\textwidth}{@{\extracolsep{\fill}}lcccccc@{}}
\toprule
Model & \makecell{FEV\\MASE} & \makecell{FEV\\SQL} & \makecell{TIME\\MASE} & \makecell{TIME\\CRPS} & \makecell{GIFT\\MASE} & \makecell{GIFT\\CRPS} \\
\midrule
Timer-M1 & \textbf{\textcolor{red}{0.3771}} & \textbf{\textcolor{red}{0.4887}} & \textbf{\textcolor{red}{0.6372}} & \textbf{\textcolor{red}{0.5363}} & \secondres{0.6806} & \secondres{0.4688} \\
TimesFM-3 & \secondres{0.3742} & \secondres{0.4866} & \secondres{0.6398} & \secondres{0.5363} & \textbf{\textcolor{red}{0.6668}} & \textbf{\textcolor{red}{0.4557}} \\
Chronos-2 & 0.3550 & 0.4728 & 0.6620 & 0.5563 & 0.6978 & 0.4854 \\
TiRex-2 & 0.3374 & 0.4550 & -- & -- & 0.6973 & 0.4781 \\
Toto 2.0 (2.5B) & 0.3254 & 0.4442 & 0.6419 & 0.5394 & 0.6956 & 0.4759 \\
TimesFM-2.5 & 0.3562 & 0.4668 & 0.6686 & 0.5674 & 0.7050 & 0.4903 \\
Timer-S1 & -- & -- & 0.7021 & 0.5886 & 0.6934 & 0.4853 \\
Sundial & 0.2522 & 0.3387 & 0.7576 & 0.6634 & 0.7502 & 0.5590 \\
Moirai 2.0 & 0.2803 & 0.4025 & 0.7030 & 0.5885 & 0.7281 & 0.5164 \\
TiRex & 0.3007 & 0.4268 & 0.6831 & 0.5731 & 0.7158 & 0.4885 \\
Chronos-Bolt & 0.2652 & 0.3889 & 0.7328 & 0.6200 & 0.8076 & 0.5743 \\
\bottomrule
\end{tabular*}
\end{table}

FEV evaluates heterogeneous variate roles, TIME emphasizes task-specific forecasting configurations, and GIFT-Eval spans diverse temporal resolutions and forecast lengths. Reporting the metrics separately avoids combining point and distributional accuracy into an arbitrary composite score.

\clearpage
\subsection{FEV Task Types}
The five groups partition all FEV configurations. Univariate and multivariate groups have no dynamic covariates; the remaining groups distinguish past-only inputs, known-future inputs, and their combination. Covariate tasks may contain multiple target variates. Each group uses the same geometric aggregation as the overall benchmark, rather than an arithmetic average of the displayed skill values.
\begin{table}[htbp]
\centering
\caption{FEV MASE skill by task type. Groups are mutually exclusive and include all 100 configurations; higher is better. Best and second-best values are marked within each row.}
\label{tab:fev_mase}
\small
\renewcommand{\arraystretch}{1.18}
\setlength{\tabcolsep}{3pt}
\begin{tabular*}{\textwidth}{@{\extracolsep{\fill}}lcccccc@{}}
\toprule
Task type & $n$ & Timer-M1 & TimesFM-3 & Chronos-2 & TiRex-2 & Toto 2.0 \\
\midrule
Univariate & 32 & \secondres{0.2968} & \textbf{\textcolor{red}{0.2998}} & 0.2799 & 0.2629 & 0.2653 \\
Multivariate & 26 & \secondres{0.4266} & 0.4252 & 0.4017 & 0.3889 & \textbf{\textcolor{red}{0.4366}} \\
Past-only & 12 & \textbf{\textcolor{red}{0.4256}} & 0.3933 & 0.3655 & \secondres{0.3969} & 0.3426 \\
Known-future & 18 & \secondres{0.4481} & \textbf{\textcolor{red}{0.4549}} & 0.4323 & 0.3979 & 0.3332 \\
Known + past & 12 & \secondres{0.2988} & 0.2923 & \textbf{\textcolor{red}{0.3029}} & 0.2465 & 0.1710 \\
\bottomrule
\end{tabular*}
\end{table}
\begin{table}[htbp]
\centering
\caption{FEV SQL skill by task type. Groups are mutually exclusive and include all 100 configurations; higher is better. Best and second-best values are marked within each row.}
\label{tab:fev_sql}
\small
\renewcommand{\arraystretch}{1.18}
\setlength{\tabcolsep}{3pt}
\begin{tabular*}{\textwidth}{@{\extracolsep{\fill}}lcccccc@{}}
\toprule
Task type & $n$ & Timer-M1 & TimesFM-3 & Chronos-2 & TiRex-2 & Toto 2.0 \\
\midrule
Univariate & 32 & \secondres{0.3828} & \textbf{\textcolor{red}{0.3847}} & 0.3703 & 0.3538 & 0.3563 \\
Multivariate & 26 & \secondres{0.5954} & 0.5953 & 0.5795 & 0.5668 & \textbf{\textcolor{red}{0.6033}} \\
Past-only & 12 & \textbf{\textcolor{red}{0.4668}} & 0.4382 & 0.4239 & \secondres{0.4387} & 0.3950 \\
Known-future & 18 & \secondres{0.5363} & \textbf{\textcolor{red}{0.5458}} & 0.5257 & 0.4971 & 0.4344 \\
Known + past & 12 & \textbf{\textcolor{red}{0.4295}} & 0.4171 & \secondres{0.4254} & 0.3765 & 0.3022 \\
\bottomrule
\end{tabular*}
\end{table}

The breakdown retains both point and probabilistic scores to distinguish temporal forecasting from the use of additional cross-variate information. In particular, separating past-only and known-future inputs distinguishes inference from related history from conditioning on observations available over the forecast interval. This organization follows the variate roles used to construct forecasting episodes, while keeping the evaluation tasks fixed across models. Overall FEV skill is recomputed from all configuration-level errors, not from an unweighted mean of the five group scores, because group sizes differ.

\clearpage
\subsection{Primitive Groups}
\label{sec:diagnostic_details}
Tables~\ref{tab:primitive_temporal}--\ref{tab:primitive_reliability} expand the groups displayed in Figure~\ref{fig:primitives}. They retain every record within each displayed group and all available compared models. These groups are a selected view of the diagnostic suite, not its exhaustive inventory; no overall rank over the entire suite is inferred from them. Window lengths, observation conditions, and metric definitions are specified in Appendix~\ref{sec:primitive_protocol}. The count $n$ denotes scoring records, which may include multiple views of the same generated task.
\begin{table}[htbp]
\centering
\caption{Temporal extrapolation groups corresponding to Figure~\ref{fig:primitives}. Mean squared error is averaged over the complete records of each displayed group; lower is better.}
\label{tab:primitive_temporal}
\small
\renewcommand{\arraystretch}{1.18}
\setlength{\tabcolsep}{3pt}
\begin{tabular*}{\textwidth}{@{\extracolsep{\fill}}lcccccc@{}}
\toprule
Group & $n$ & Timer-M1 & TimesFM-3 & Chronos-2 & TiRex-2 & Toto 2.0 \\
\midrule
Shift & 6 & \textbf{\textcolor{red}{6.4278}} & 7.4830 & 6.8490 & 6.9092 & \secondres{6.4418} \\
Periodic & 13 & \textbf{\textcolor{red}{0.0271}} & 0.6803 & 0.0919 & \secondres{0.0735} & 0.2300 \\
Trend & 10 & 0.1312 & \secondres{0.0299} & 0.4522 & 0.4512 & \textbf{\textcolor{red}{0.0083}} \\
\bottomrule
\end{tabular*}
\end{table}
\begin{table}[htbp]
\centering
\caption{Relational forecasting groups corresponding to Figure~\ref{fig:primitives}. Mean squared error is averaged over the complete records of each displayed group; lower is better.}
\label{tab:primitive_relations}
\small
\renewcommand{\arraystretch}{1.18}
\setlength{\tabcolsep}{3pt}
\begin{tabular*}{\textwidth}{@{\extracolsep{\fill}}lcccccc@{}}
\toprule
Group & $n$ & Timer-M1 & TimesFM-3 & Chronos-2 & TiRex-2 & Toto 2.0 \\
\midrule
Sign & 2 & \textbf{\textcolor{red}{0.0021}} & \secondres{0.0036} & 0.0077 & 0.0100 & 0.0073 \\
Cointegration & 2 & \textbf{\textcolor{red}{0.2477}} & \secondres{0.3953} & 2.2627 & 1.3930 & 4.7848 \\
Promotion & 2 & \textbf{\textcolor{red}{0.3008}} & \secondres{0.3042} & 1.0424 & 0.5567 & -- \\
Lag & 2 & \secondres{0.0012} & \textbf{\textcolor{red}{0.0005}} & 0.0185 & 0.0074 & 0.0049 \\
Phase & 2 & \secondres{0.000053} & \textbf{\textcolor{red}{0.000043}} & 0.0002 & 0.0087 & 0.0010 \\
Scale & 2 & \textbf{\textcolor{red}{0.0688}} & \secondres{0.0955} & 0.8690 & 2.2971 & 4.7189 \\
Shift & 2 & \textbf{\textcolor{red}{0.1060}} & \secondres{0.2068} & 0.2593 & 0.3129 & 0.4098 \\
\bottomrule
\end{tabular*}
\end{table}
\begin{table}[htbp]
\centering
\caption{Input reliability groups corresponding to Figure~\ref{fig:primitives}. Mean squared error is averaged over the complete records of each displayed group; lower is better.}
\label{tab:primitive_reliability}
\small
\renewcommand{\arraystretch}{1.18}
\setlength{\tabcolsep}{3pt}
\begin{tabular*}{\textwidth}{@{\extracolsep{\fill}}lcccccc@{}}
\toprule
Group & $n$ & Timer-M1 & TimesFM-3 & Chronos-2 & TiRex-2 & Toto 2.0 \\
\midrule
Irrelevant context & 7 & 0.3681 & 0.2588 & \secondres{0.2554} & 0.2585 & \textbf{\textcolor{red}{0.2451}} \\
Irrelevant targets & 3 & \textbf{\textcolor{red}{0.0035}} & -- & 0.0167 & 0.0633 & \secondres{0.0055} \\
Missing event inputs & 7 & 5.7917 & 5.7594 & \secondres{3.9130} & \textbf{\textcolor{red}{2.2120}} & -- \\
Target corruption & 21 & 12.0938 & \secondres{2.6081} & 23.9802 & \textbf{\textcolor{red}{0.2260}} & 3.8370 \\
Target missingness & 21 & \secondres{0.1957} & \textbf{\textcolor{red}{0.1477}} & 0.7231 & 0.2184 & 0.3595 \\
Variate order & 7 & \textbf{\textcolor{red}{0.0034}} & -- & 0.0221 & 0.0623 & \secondres{0.0055} \\
\bottomrule
\end{tabular*}
\end{table}

Dashes denote unsupported inputs or unavailable squared-error results, not failed forecasts assigned an artificial penalty. TimesFM-3 uses the fixed configuration for these controlled tests; public-benchmark scores follow their own evaluation settings. Selected examples in Appendix~\ref{sec:showcases} illustrate the observed temporal patterns and relations without changing the numerical groups above.

\clearpage
\subsection{Ablation Results}
\label{sec:ablation_protocol}
Gate controls are independently trained with fixed residual multipliers, retaining variate attention and the same residual order. Fixed 0.202 uses the unrounded stable-reference mean, approximately $0.201676$, while Fixed 1.0 uses the full variate-attention update. Adaptive gates are learned separately at each layer and remain fixed during inference. These comparisons are not inference-only interventions on a common checkpoint.
\begin{table}[htbp]
\centering
\caption{Gate ablation on FEV overall and five task types. Both metrics report skill scores. Red bold and underline indicate the best and second-best results within each column and metric. Adaptive gates achieve the highest overall skill, with the clearest advantage in known-future forecasting.}
\label{tab:gate_benchmarks}
\small
\renewcommand{\arraystretch}{1.18}
\setlength{\tabcolsep}{3pt}
\begin{tabular*}{\textwidth}{@{\extracolsep{\fill}}lcccccc@{}}
\toprule
Gate & Overall & Univariate & Multivariate & Past-only & Known-future & Known + past \\
\midrule
\multicolumn{7}{c}{MASE Skill $\uparrow$} \\
\midrule
Adaptive
& \textcolor{red}{\textbf{0.3771}}
& \underline{0.2968}
& \underline{0.4266}
& \underline{0.4256}
& \textcolor{red}{\textbf{0.4481}}
& 0.2988 \\
Fixed 0.202
& \underline{0.3714}
& \textcolor{red}{\textbf{0.2969}}
& \textcolor{red}{\textbf{0.4299}}
& \textcolor{red}{\textbf{0.4266}}
& \underline{0.4084}
& \underline{0.3081} \\
Fixed 1.0
& 0.3680
& 0.2963
& 0.4260
& 0.4238
& 0.3979
& \textcolor{red}{\textbf{0.3104}} \\
\midrule
\multicolumn{7}{c}{SQL Skill $\uparrow$} \\
\midrule
Adaptive
& \textcolor{red}{\textbf{0.4887}}
& \textcolor{red}{\textbf{0.3828}}
& 0.5954
& \textcolor{red}{\textbf{0.4668}}
& \textcolor{red}{\textbf{0.5363}}
& 0.4295 \\
Fixed 0.202
& \underline{0.4836}
& \underline{0.3826}
& \textcolor{red}{\textbf{0.5981}}
& 0.4651
& \underline{0.5002}
& \underline{0.4406} \\
Fixed 1.0
& 0.4814
& 0.3819
& \underline{0.5956}
& \underline{0.4659}
& 0.4920
& \textcolor{red}{\textbf{0.4426}} \\
\bottomrule
\end{tabular*}
\end{table}

\begin{table}[htbp]
\centering
\caption{Data-construction ablations across three benchmarks. Recipe names refer to the source-generation routes under comparison; shared benchmark-associated training data are unchanged across variants.}
\label{tab:data_full}
\small
\renewcommand{\arraystretch}{1.18}
\setlength{\tabcolsep}{3pt}
\begin{tabular*}{\textwidth}{@{\extracolsep{\fill}}lcccccc@{}}
\toprule
Recipe & \makecell{FEV\\MASE $\uparrow$} & \makecell{FEV\\SQL $\uparrow$} & \makecell{TIME\\MASE $\downarrow$} & \makecell{TIME\\CRPS $\downarrow$} & \makecell{GIFT\\MASE $\downarrow$} & \makecell{GIFT\\CRPS $\downarrow$} \\
\midrule
Real Only & 0.3196 & 0.4431 & 0.6579 & 0.5526 & 0.6998 & 0.4850 \\
Synthetic Only & 0.3568 & 0.4710 & 0.6438 & 0.5456 & 0.6884 & 0.4741 \\
No-Joint & \secondres{0.3578} & \secondres{0.4725} & \secondres{0.6379} & \textbf{\textcolor{red}{0.5359}} & \secondres{0.6819} & \secondres{0.4722} \\
Full Recipe & \textbf{\textcolor{red}{0.3771}} & \textbf{\textcolor{red}{0.4887}} & \textbf{\textcolor{red}{0.6372}} & \secondres{0.5363} & \textbf{\textcolor{red}{0.6806}} & \textbf{\textcolor{red}{0.4688}} \\
\bottomrule
\end{tabular*}
\end{table}

Real Only and Synthetic Only restrict the source-generation routes under comparison; No-Joint retains both source types but removes corpus-derived and mixed-source joint generation. Shared GIFT-Eval and TIME training collections remain unchanged across the four variants. Consequently, these names describe the controlled routes rather than the exclusive provenance of all training observations. The recipe comparisons assess inclusion of generation routes and do not isolate each individual operator.

\clearpage
\section{Primitive Evaluation Showcase}
\label{sec:showcases}
The following pages present 24 illustrative forecasts, eight for each capability. Examples cover distinct temporal patterns, cross-variate relations, and observation conditions, with an emphasis on mechanisms discussed in the main text. They are selected examples rather than a random sample for estimating overall success rates. All model curves within a panel use the same target and forecast origin. The dashed boundary separates history from the forecast interval; shading denotes Timer-M1's marginal 10--90\% prediction quantiles, not confidence intervals for its mean prediction.

\begin{figure}[htbp]
\centering
\includegraphics[width=\linewidth]{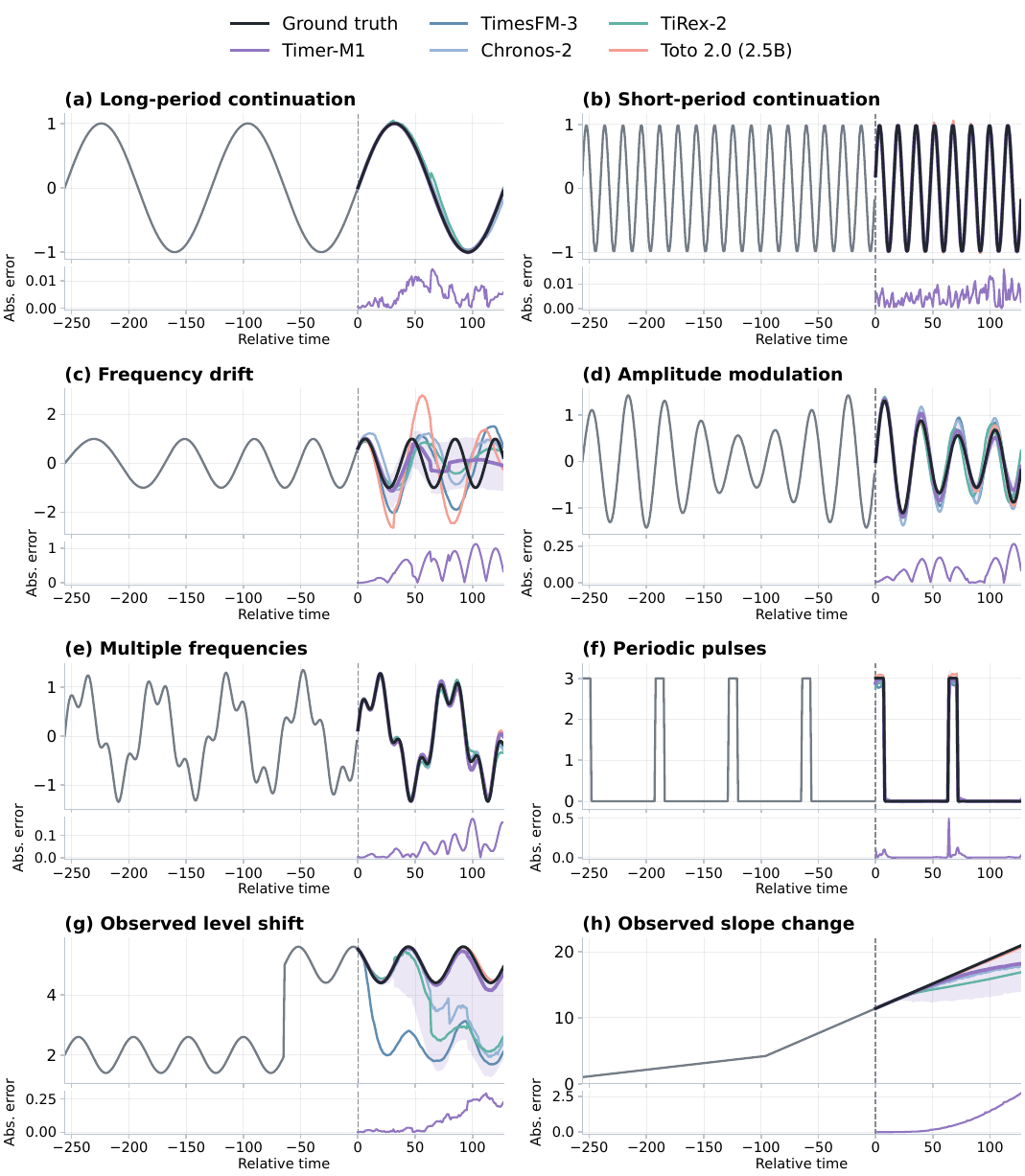}
\caption{Temporal extrapolation across periodicity, modulation, frequency drift, and observed regime changes. Lower strips show Timer-M1's absolute forecast error. Each panel uses its own value scale.}
\label{fig:temporal_showcases}
\end{figure}
\clearpage
\begin{figure}[p]
\centering
\includegraphics[width=\linewidth]{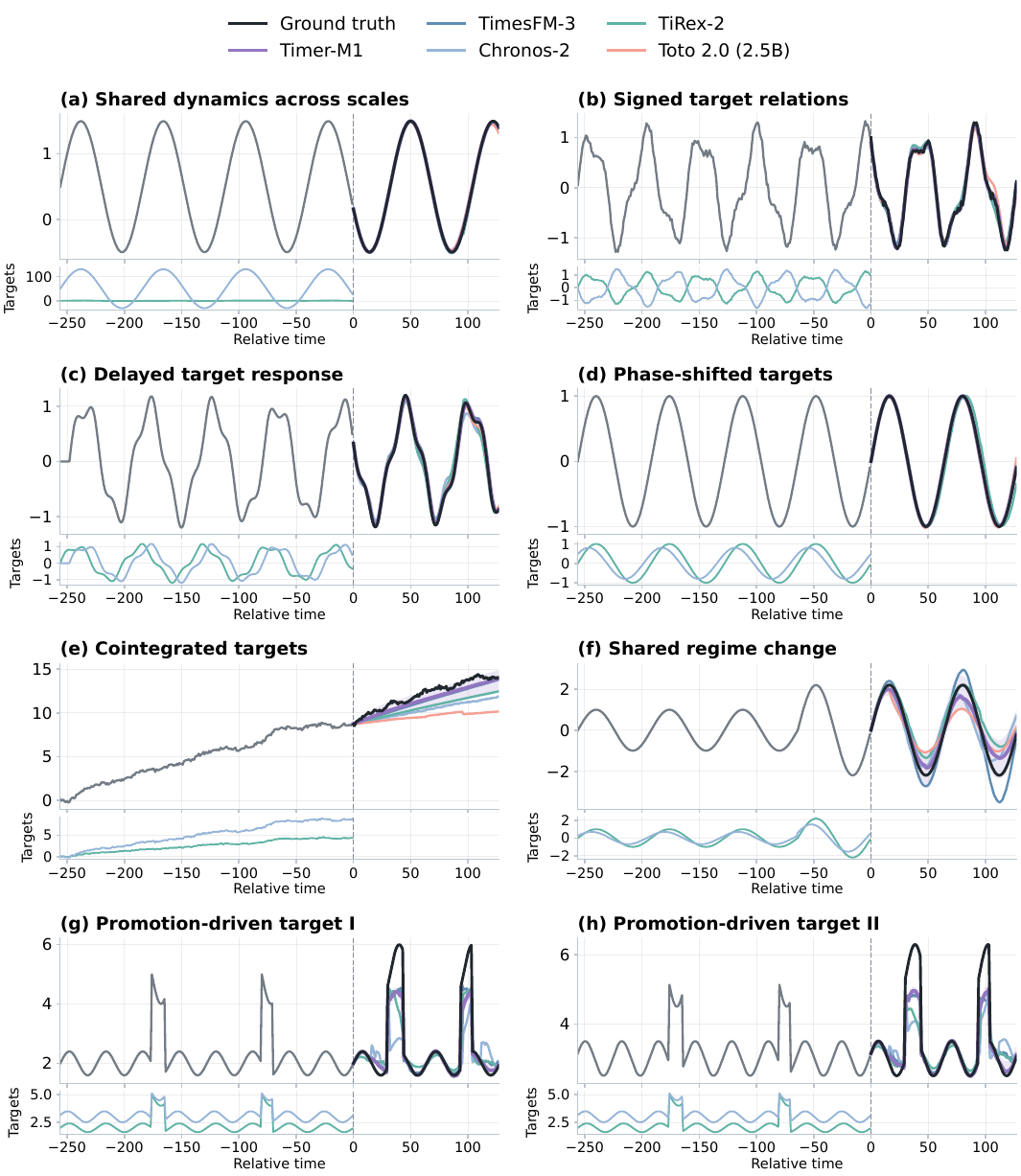}
\caption{Relational forecasting with shared, signed, delayed, phase-shifted, cointegrated, and event-driven dynamics. Lower strips show available related inputs or target histories; they are not additional forecast targets unless specified by the task.}
\label{fig:relational_showcases}
\end{figure}
\clearpage
\begin{figure}[p]
\centering
\includegraphics[width=\linewidth]{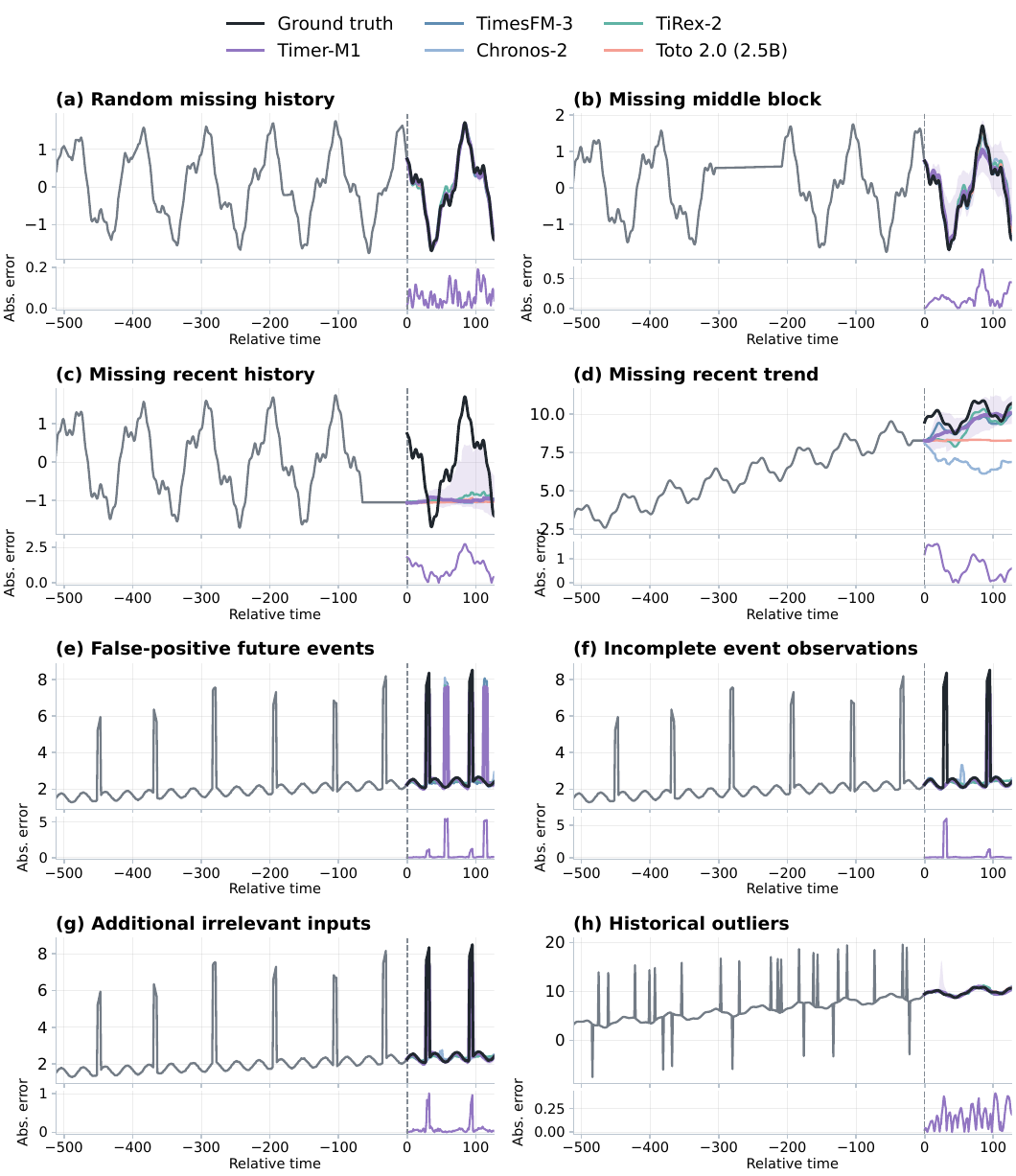}
\caption{Input reliability under missing histories, imperfect event information, irrelevant inputs, and historical outliers. Lower strips show Timer-M1's absolute forecast error.}
\label{fig:reliability_showcases}
\end{figure}
\clearpage
\end{document}